\pdfoutput=1

\documentclass[11pt]{article}

\usepackage[final]{acl}

\usepackage{times}
\usepackage{latexsym}

\usepackage[T1]{fontenc}

\usepackage[utf8]{inputenc}

\usepackage{microtype}

\usepackage{inconsolata}

\usepackage{graphicx}
\usepackage{inconsolata}
\usepackage{amsmath} 
\usepackage{tabularx}
\usepackage{arydshln}
\usepackage{enumitem}
\usepackage{comment}

\usepackage{algorithmic,algorithm}
\renewcommand{\algorithmiccomment}[1]{\bgroup\hfill$\triangleright$~#1\egroup}
\usepackage[linguistics]{forest} 
\usepackage{synttree}
\usepackage{tikz-dependency}
\usepackage{caption}
\usepackage{subcaption}
\usepackage{tablefootnote}

\usepackage{arydshln}

\usepackage[cjk]{kotex}
\usepackage{CJKutf8}
\newcommand{\zh}[1]{\begin{CJK}{UTF8}{gbsn}#1\end{CJK}}

\usepackage{langsci-gb4e}

\title{Multilingual Grammatical Error Annotation: Combining Language-Agnostic Framework with Language-Specific Flexibility}

\author{
Mengyang Qiu$^{1,2}$~~~ Tran Minh Nguyen$^{2}$~~~ Zihao Huang$^{2}$~~~ Zelong Li$^{3}$~~~ Yang Gu$^{2}$~~~ \\
\textbf{Qingyu Gao}$^{2}$~~~
\textbf{Siliang Liu}$^{2}$~~~
\textbf{Jungyeul Park}$^{2,3}$\\
$^{1}$Trent University, Canada~~~~
$^{2}$Open Writing Evaluation, France\\
$^{3}$University College London, UK~~~~
$^{4}$The University of British Columbia, Canada\\
{\tt \url{http://open-writing-evaluation.github.io}}
}

\begin{document}
\maketitle
\begin{abstract}
Grammatical Error Correction (GEC) relies on accurate error annotation and evaluation, yet existing frameworks, such as \texttt{errant}, face limitations when extended to typologically diverse languages. In this paper, we introduce a standardized, modular framework for multilingual grammatical error annotation. Our approach combines a language-agnostic foundation with structured language-specific extensions, enabling both consistency and flexibility across languages. We reimplement \texttt{errant} using \texttt{stanza} to support broader multilingual coverage, and demonstrate the framework's adaptability through applications to English, German, Czech, Korean, and Chinese, ranging from general-purpose annotation to more customized linguistic refinements. This work supports scalable and interpretable GEC annotation across languages and promotes more consistent evaluation in multilingual settings. The complete codebase and annotation tools can be accessed at \url{https://github.com/open-writing-evaluation/jp_errant_bea}.
\end{abstract}

\section{Introduction}

Grammatical Error Correction (GEC), which aims to automatically detect and correct errors in written text, has emerged as one of the most important and widely studied tasks in Natural Language Processing (NLP) for educational applications, particularly those supporting language learning and writing improvement. It benefits both native speakers (L1), by enhancing clarity and fluency in their writing, and non-native learners (L2), by providing immediate, structured feedback that reinforces correct grammatical patterns, boosts writing confidence, and, ultimately, supports language development and acquisition \citep{marjokorpi-2023-relationship,vanbeuningen-etal-2012-evidence}. Over the years, the lion's share of research has focused on advancing GEC systems—evolving from rule-based and statistical approaches to neural architectures, such as neural machine translation with transformers \citep{zhao-EtAl:2019:NAACL} and, more recently, prompting-based approaches built on large language models (\citealp{zeng-etal-2024-evaluating-prompting}; for a comprehensive review, see \citealp{bryant-etal-2023-gec}). 

Yet, automatic error annotation and evaluation play an equally critical role in GEC. Error annotation identifies and categorizes linguistic errors, while evaluation measures how effectively GEC systems correct them. Together, these two components help establish standardized benchmarks, influencing everything from system development to the quality of corrections eventually delivered to users. However, despite their importance, they have historically received less attention and are often treated as ancillary to system development and dataset creation.

Among existing tools for automatic error annotation and evaluation (e.g., \texttt{M$^2$}, \citealp{dahlmeier-ng-2012-better}; \texttt{GLEU}, \citealp{napoles-etal-2015-ground}), \texttt{errant} ({ERR}or {AN}notation {T}oolkit) has established itself as the \textit{de facto} framework for English GEC. What makes \texttt{errant} stand out is its detailed linguistic annotations, with a total of 55 possible error types for English \citep{bryant-etal-2017-automatic}. \texttt{errant}'s significance was solidified in the \textit{Building Educational Applications 2019 Shared Task: Grammatical Error Correction (BEA-2019)}, where it was used to standardize multiple datasets and served as the official scorer \citep{bryant-etal-2019-bea}. 

While this toolkit has proven effective for English, further refinements are needed to improve its versatility and adaptability, especially in multilingual scenarios. Recent years have seen growing interest in multilingual GEC, as demonstrated by initiatives like the \textit{MultiGEC-2025 Shared Task}, which brought together efforts across twelve typologically diverse European languages \citep{masciolini-etal-2025-multigec,masciolini-etal-2025-towards}. However, this surge in interest has outpaced the development of consistent multilingual annotation resources. 

As noted by \citet{masciolini-etal-2025-multigec}, only three languages in MultiGEC--namely Czech, German, and Greek--have received \texttt{errant}-style annotation. For the remaining languages, the authors acknowledge that, due to limited time and resources, they implemented only coarse-grained alignment between original and corrected texts to support holistic scoring, without access to the kind of detailed error analysis enabled by \texttt{errant} for English. Even in existing adaptations of \texttt{errant} for various languages, implementations vary considerably in their design choices--ranging from annotation label schemes to tokenization and part-of-speech (POS) tagging tools--and differ in the level of granularity applied to language-specific error types. Although differences in orthographies and morphosyntactic structures across languages are unavoidable, greater consistency in annotation practices is highly desirable.

To address these challenges, our goal is to develop a consistent and reusable framework for grammatical error annotation that can be readily adapted across typologically diverse languages. Drawing inspiration from the original \texttt{errant}'s dataset-agnostic design, we extend its core philosophy to multilingual settings by separating the annotation pipeline into two components: a shared architecture that applies across languages, and optional extensions tailored to language-specific features. Even within the language-specific layer, we introduce structured templates for common error types, such as spelling, word order, and word boundary errors, which can be reused or adapted across languages with similar orthographic or syntactic patterns. In addition, our implementation relies on the \texttt{stanza} toolkit for tokenization and POS tagging, which provides standardized processing pipelines for over 70 languages \citep{qi-etal-2020-stanza}, allowing our framework to be readily extended to annotate new GEC datasets of other languages when they become available.
 
The rest of the paper is organized as follows: $\S$\ref{background} reviews the original \texttt{errant} framework, discusses challenges in its multilingual adaptations, and motivates the use of \texttt{stanza} for more consistent cross-linguistic preprocessing. $\S$\ref{error_typology} introduces our proposed grammatical error typology, which combines a language-agnostic core with structured, language-specific extensions. $\S$\ref{implementation} presents our reimplementation of English \texttt{errant} and demonstrates the framework's applicability to multiple languages, ranging from generic use in European languages, to minor template refinements for Korean, and deeper customization for Chinese. Finally, $\S$\ref{conclusion} summarizes our contributions and emphasizes the framework's flexibility and extensibility for multilingual GEC.

\section{Background and Related Work}\label{background}

\subsection{Description of \texttt{errant}}

\texttt{errant} is a unified framework for error annotation and evaluation in English GEC. It provides a rule-based, dataset-agnostic approach for extracting and categorizing edits between original and corrected sentences, making it a crucial tool for system evaluation and benchmarking \citep{bryant-etal-2017-automatic}.

At the core of its annotation pipeline is a linguistically enhanced alignment algorithm that identifies edit boundaries between sentence pairs. This algorithm, originally proposed by \citet{felice-bryant-briscoe:2016:COLING}, extends the Damerau-Levenshtein distance with a linguistically informed cost function that considers part-of-speech tags, lemmas, and character similarity. Unlike surface-level edit distance, this method prioritizes alignments between tokens that are syntactically or morphologically related (e.g., \textit{meet} and \textit{meeting}), and handles both one-to-one edits and multi-token reordering. A rule-based merging strategy is then applied to combine adjacent edits where appropriate, based on patterns frequently observed in learner data, such as phrasal verb edits. This alignment process significantly improves the consistency and quality of extracted edits \citep{felice-bryant-briscoe:2016:COLING}.

Following alignment, \texttt{errant} applies a rule-based annotation scheme to categorize edits into fine-grained grammatical error types, enabling both comprehensive feedback and error-type evaluation. Specifically, it defines 25 primary error types based on POS and morphological properties obtained from \texttt{spaCy}\footnote{\url{https://spacy.io}}, and further classifies them into three edit operations: \texttt{M}issing, \texttt{U}nnecessary, and \texttt{R}eplacement, resulting in a total of 55 possible error types (e.g., \texttt{R:VERB:TENSE} indicates a replacement error related to verb tense). To store annotations, \texttt{errant} generates output in \texttt{M2} format, the standard representation for GEC annotations since its adoption in the \textit{CoNLL-2013 Shared Task} \citep{ng-EtAl:2013:CoNLL}. Each annotated sentence consists of the original tokenized text (denoted by an \textit{S} line) followed by one or more error annotation lines (\textit{A} lines). Each \textit{A} line specifies the error span, the error type, the suggested correction, and additional metadata (see Figure~\ref{fig:m2_example} for an example in English).

\begin{figure}[!ht]
\centering
\resizebox{\columnwidth}{!}{%
\footnotesize
\begin{tabular}{l}
\hline
\texttt{S This are a sentence .} \\  
\texttt{A 1 2|||R:VERB:SVA|||is|||-REQUIRED-|||NONE|||0} \\  
\texttt{A 3 3|||M:ADJ|||good|||-REQUIRED-|||NONE|||0} \\  
\hline
\end{tabular}
}%
\caption{Example of an annotated sentence in \texttt{M2} format from \textit{BEA-2019}.}
\label{fig:m2_example}
\end{figure}

With edits extracted and categorized in a standardized format, \texttt{errant} can then be used to systematically evaluate GEC system outputs against gold-standard references. It calculates precision and recall between system-generated edits and gold-standard corrections and utilizes a harmonic mean $F_{0.5}$ score, which weights precision twice as much as recall to prioritize accurate and contextually appropriate corrections over excessive edits. Thanks to its detailed annotation schema, \texttt{errant} supports multi-granularity evaluation--analyzing system effectiveness not only at the overall level but also across specific error types and edit operations, enabling a fine-grained and transparent assessment of GEC models.

In \textit{BEA-2019}, \texttt{errant} was used to standardize multiple datasets, some of which were annotated using different error type frameworks, while others lacked annotations entirely. This allowed for error distribution comparisons across datasets that were previously hindered by these annotation discrepancies. In addition, \texttt{errant} facilitated multi-level system evaluation by supporting error-type analysis across 24 main categories for all 21 participating teams\footnote{\texttt{errant} defines 25 categories, including \texttt{UNK}nown (error detected but unable to be corrected; \citealp{bryant-etal-2017-automatic}). In \textit{BEA-2019}, this category was not included.}. This enabled a detailed assessment of each system’s strengths and weaknesses and made it easier to identify which error types were the most challenging to correct \citep{bryant-etal-2019-bea}. 

While \texttt{errant} provides a linguistically informed foundation for GEC annotation and evaluation, it is not without limitations. One minor issue is its tendency to overuse the \texttt{OTHER} category (i.e., unspecified errors), leading to less precise error categorization. For instance, certain errors that could be classified as specific grammatical types (e.g., verb tense or prepositions) are instead grouped under \texttt{OTHER} \citep{korre-pavlopoulos-2020-errant}. 

Another issue, as discussed in \citet{wang-etal-2025-refined}, arises in end-to-end evaluation scenarios. \texttt{errant} assumes pre-defined sentence boundaries, and misalignment can result in an inability to generate evaluation results between gold-standard references and system outputs. However, in real-world GEC applications, such as learner essays, inconsistencies in sentence segmentation are a common issue, often caused by differences in preprocessing steps. To address this, \citet{wang-etal-2025-refined} introduced joint-preprocessing \texttt{errant}, incorporating an alignment-based approach to detect and resolve segmentation discrepancies before evaluation. 

\subsection{Challenges in existing multilingual adaptations of \texttt{errant}}

Given its demonstrated success in English, \texttt{errant} has been adapted to multiple languages, including Arabic \citep{belkebir-habash-2021-automatic}, Chinese \citep{hinson-etal-2020-heterogeneous,zhang-etal-2022-mucgec,gu-etal-2025-improving}, Czech \citep{naplava-etal-2022-czech}, German \citep{boyd-2018-using}, Greek \citep{korre-etal-2021-elerrant}, Hindi \citep{sonawane-etal-2020-generating}, and Korean \citep{yoon-etal-2023-towards}. While these adaptations have enabled broader use of \texttt{errant}-style annotation, they also reveal several challenges that arise when extending the framework to languages with a range of orthographic and morphosyntactic characteristics.

\paragraph{Inconsistent annotation labels}

A minor issue in multilingual adaptations of \texttt{errant} is inconsistent annotation labels for similar error types. The original \texttt{errant} for English defines three edit operations: \texttt{M}issing, \texttt{U}nnecessary, and \texttt{R}eplacement, while treating word order (\texttt{WO}) as a main error category, similar to \texttt{NOUN} or \texttt{VERB} errors. In \texttt{errant\_zh}, an adaptation for Chinese, these operations were denoted as insertion, deletion, substitution, and transposition \citep{hinson-etal-2020-heterogeneous}. Meanwhile, \texttt{ChERRANT}, another Chinese adaptation, later revised them to \texttt{M}issing, \texttt{R}edundant, \texttt{S}ubstitute, and \texttt{W}ord-order \citep{zhang-etal-2022-mucgec}. While these differences do not affect core functionality, the lack of consistent labeling across adaptations can create confusion. Nevertheless, this issue is relatively straightforward to address, as it primarily involves terminology standardization.

\paragraph{Inconsistent preprocessing tools}

A moderate challenge in multilingual GEC annotation lies in linguistic preprocessing tools, particularly word segmentation and POS tagging. While \texttt{spaCy}, the default NLP library in \texttt{errant} for English, supports multiple languages, its effectiveness varies across linguistic systems, prompting many adaptations to incorporate alternative tools. For example, German \texttt{errant} retained much of the \texttt{spaCy} pipeline but found its lemmatization insufficient, replacing it with TreeTagger for better accuracy \citep{boyd-2018-using}. For non-European languages, entirely different tools are used, such as Kkma POS Tagger for Korean \texttt{KAGAS} \citep{yoon-etal-2023-towards} and LTP (Language Technology Platform) for Chinese \texttt{ChERRANT} \citep{zhang-etal-2022-mucgec}. 

Although these variations allow for language-specific optimizations, different tokenization strategies and POS tagging schemes can lead to discrepancies in how errors are identified and classified. This is particularly problematic for multilingual GEC models, where standardized evaluation across multiple languages is crucial. Since a system's measured performance is inherently tied to how its errors are annotated, such variations can obscure true system similarities or differences and compromise the reliability of multilingual benchmarks.

\paragraph{Inconsistent annotation granularity}

A more significant challenge in multilingual GEC annotation involves the varying levels of granularity for language-specific errors. While \texttt{errant} provides detailed error categories for English, adaptations to other languages, especially non-European languages, often fail to maintain this level of detail. For instance, \texttt{errant\_zh} uses only four basic edit operations at the character level, without POS information \citep{hinson-etal-2020-heterogeneous}. 

Recent work has begun addressing this limitation by introducing more fine-grained annotations tailored to specific linguistic properties. For example, \citet{gu-etal-2025-improving} propose a refined error typology for Chinese that accounts for phonetic similarity, visual similarity, and other structural errors specific to Chinese. While this framework was developed for Chinese, many of its principles can be readily applied to languages with similar logographic orthographies.

\subsection{\texttt{stanza} as a multilingual alternative to \texttt{spaCy}}\label{background:stanza}


The original \texttt{errant} framework relies on \texttt{spaCy} for preprocessing tasks such as tokenization and POS tagging. However, \texttt{spaCy}'s multilingual capabilities are relatively limited, covering only a small number of languages and exhibiting inconsistent performance across linguistic families. This has contributed to the fragmented landscape of language-specific adaptations in prior \texttt{errant} variants.

To promote cross-lingual consistency, our implementation adopts \texttt{stanza} \citep{qi-etal-2020-stanza}, a fully neural pipeline trained on Universal Dependencies (UD) and other multilingual corpora. \texttt{stanza} supports over 70 languages and applies a consistent architecture and UD-based annotation scheme across its modules—including tokenization, multi-word token expansion, POS and morphological tagging, dependency parsing, and named entity recognition\footnote{\url{https://stanfordnlp.github.io/stanza/}}. Benchmark evaluations indicate strong performance across typologically diverse languages.

Crucially, our aim is not to promote a specific tool, but to align the preprocessing stage with the same linguistic principles that underlie our error typology. Like UD, our taxonomy adopts a cross-linguistically consistent core structure with optional language-specific extensions. Using a UD-compatible parser such as \texttt{stanza} ensures that all languages are analyzed under a shared morphosyntactic framework, which is essential for scalable and comparable multilingual grammatical error annotation. In this sense, it is the UD standard, rather than any particular NLP library, that provides the conceptual and practical foundation for our approach.

\section{Multilingual Error Typology}\label{error_typology}

An error typology provides a systematic framework for identifying, classifying, and analyzing errors in written text. We propose a two-tiered typology consisting of a language-agnostic foundation and a set of structured, language-specific extensions. The first level includes the widely adopted \texttt{MRU} (\texttt{M}issing, \texttt{R}eplacement, \texttt{U}nnecessary) framework, ensuring consistency in annotation and evaluation across linguistic systems. The second level provides a structured template for language-specific extensions, allowing related languages to share annotation strategies and avoid redundant reimplementation. By designing this layered approach, we promote standardization across languages while allowing flexibility for language-specific refinements.

\subsection{Language-agnostic error annotation}

The \texttt{MRU} framework classifies errors into three core operations: \texttt{M}issing (\texttt{M}), where essential elements are omitted; \texttt{R}eplacement (\texttt{R}), where an incorrect element substitutes the correct one; and \texttt{U}nnecessary (\texttt{U}), where superfluous elements cause redundancy. Each error is further specified with POS tags for precise categorization. 

\paragraph{\texttt{M}issing (\texttt{M})}

An essential linguistic element is omitted from a sentence, leading to incomplete or ungrammatical structures. These errors typically involve the absence of words or phrases necessary for grammaticality or semantic clarity, such as missing determiners. In annotation, missing errors are further categorized based on POS tags or syntactic functions. For example, \texttt{M:NOUN} indicates a missing noun.

\paragraph{\texttt{R}eplacement (\texttt{R})}

An incorrect linguistic element is used in place of the correct one. These errors frequently involve incorrect word forms or inappropriate lexical choices (e.g., \texttt{R:VERB} denotes an erroneous verb substitution). To further reduce ambiguity in annotation, we implement the \texttt{R:P$_{1}$$\rightarrow$P$_{2}$} pattern, where \texttt{P$_{1}$} is replaced by \texttt{P$_{2}$}.

\paragraph{\texttt{U}nnecessary (\texttt{U})}

A superfluous linguistic element is present in a sentence, resulting in redundancy or ungrammaticality. These errors often involve extraneous words or phrases that disrupt sentence structure or meaning. Similar to missing and replacement errors, unnecessary errors are annotated with POS information to specify the redundant element. For example, \texttt{U:DET} denotes an unnecessary determiner.

\subsection{Language-specific error annotation}
    
To accommodate language-specific characteristics, we introduce a set of structured extensions to the \texttt{MRU} core. Our approach maintains consistency with established annotation schemes such as \texttt{errant} while capturing morphological and syntactic errors unique to different languages.
Algorithm~\ref{error-classification-algorithm} presents our proposed classification routine for \texttt{R}eplacement errors. Given a pair of word sequences—the source (\(\mathcal{S}\)) and the target (\(\mathcal{T}\))—the algorithm classifies the error into one of the following types: spelling errors (\textsc{r:spell}), word order errors (\textsc{r:wo}), or word boundary errors (\textsc{r:wb}). Spelling similarity is computed using two metrics: phonetic similarity and visual (shape-based) similarity. The thresholds \(\alpha_1\) and \(\alpha_2\) govern sensitivity to phonetic and visual matches, respectively.

The classification uses the following notation:
\begin{itemize}
    \item \(\mathcal{S}, \mathcal{T}\): word sequences in the source and target sentences.
    \item \textsc{Sim}(\textit{phonetic}) and \textsc{Sim}(\textit{shape}): similarity functions comparing pronunciation and visual form.
    \item \textsc{Set}($\mathcal{S}$): returns a bag-of-words representation of \(\mathcal{S}\), disregarding word order.
    \item \textsc{Merge}($\mathcal{S}$): reconstructs a character sequence from the tokenized input (i.e., merging tokens without spaces) to test for boundary alignment.
\end{itemize}

This structured yet extensible framework allows consistent error categorization across languages, while also accommodating language-specific scripts and segmentation conventions.

\begin{algorithm}[!ht]
\caption{Pseudo-code for error classification}\label{error-classification-algorithm}
{\footnotesize
\begin{algorithmic}[1]
\STATE{\textbf{function} \textsc{ErrorClassification} ($\mathcal{S}$, $\mathcal{T}$):}
\begin{ALC@g}
\IF{ (\textsc{Sim} (\textit{phonetic}) $> \alpha_{1}$) $\wedge$ (\textsc{Sim} (\textit{shape}) $> \alpha_{2}$)} 
    \RETURN{ \textsc{r:spell:phonographic} }
\ELSIF{ (\textsc{Sim} (\textit{phonetic}) $> \alpha_{1}$) }
    \RETURN{ \textsc{r:spell:phonetic} }
\ELSIF{ (\textsc{Sim} (\textit{shape}) $ > \alpha_{2}$) }
    \RETURN{ \textsc{r:spell:shape} }    
\ELSIF{ (\textsc{Set} ($\mathcal{S}$) == \textsc{Set} ($\mathcal{T}$)) }
    \RETURN{ \textsc{r:wo} }
\ELSIF{ (\textsc{Merge}($\mathcal{S}$) == \textsc{Merge}($\mathcal{T}$)) }
    \RETURN{ \textsc{r:wb} }
\ENDIF
\RETURN { \{\textsc{r}\} }
\end{ALC@g}
\end{algorithmic} 
}
\end{algorithm} 

\paragraph{Spelling errors}

We classify spelling errors by their underlying cause: sound-based phonetic similarity (\textsc{r:spell:phonetic}), visual resemblance in orthographic shape (\textsc{r:spell:shape}), or a combination of both (\textsc{r:spell:phonographic}). 
For sound-based phonetic errors, we introduce a transcription system to represent pronunciation, such as a pronouncing dictionary for English, \textit{pinyin} for Chinese, or romanization for other languages. This allows us to compare words based on their phonetic similarity and identify errors caused by mispronunciation or phoneme substitution.
For orthographic shape errors, we assess visual similarity by converting characters into font images and applying similarity metrics. This approach helps detect errors caused by visually similar characters, such as mistyped letters in Latin-based scripts or miswritten strokes in logographic writing systems like Chinese and Japanese. By combining these methods, we systematically classify and analyze spelling errors across different languages.

\paragraph{Word order errors}
Word order errors are flagged when the source sequence (\(\mathcal{S}\)) and the target sequence (\(\mathcal{T}\)) contain the same set of words but differ in arrangement. In such cases, all words from the original sequence are retained, but their relative positions are altered. These errors are particularly common in languages with flexible word order, where reordering affects grammaticality or readability. Identifying and categorizing such errors enables more structured syntactic analysis and improves grammatical error correction.

\paragraph{Word boundary errors} 
Word boundary errors occur when the source sequence (\(\mathcal{S}\)) and the target sequence (\(\mathcal{T}\)) yield the same sequence after merging their respective word components. These errors typically involve incorrect spacing, where words that should remain separate are mistakenly merged, or conversely, a single word is improperly split into multiple tokens. Since the fundamental content remains unchanged but the segmentation differs, such errors impact readability, syntactic structure, and lexical integrity. Addressing these errors ensures accurate word segmentation and proper grammatical representation.

Figure~\ref{error-example} illustrates representative examples of the three major subtypes of \texttt{R}eplacement errors classified by our algorithm.

\begin{figure}[!ht]
    \centering
\resizebox{.48\textwidth}{!}{    
\footnotesize{
\begin{tabular}{rl} \hline 
 \textsc{r:spell:phonetic}& \textit{their} $\rightarrow$ \textit{there} \\
 \textsc{r:wo}& \textit{You can help me} $\rightarrow$ \textit{Can you help me} \\
 \textsc{r:wb}& \textit{ice cream} $\rightarrow$ \textit{icecream} \\ \hline
\end{tabular}}
}
\caption{Examples of \texttt{R}eplacement error types: phonetic spelling error (\textsc{r:spell:phonetic}), word order error (\textsc{r:wo}), and word boundary error (\textsc{r:wb}).}

\label{error-example}
\end{figure}

\section{Implementation of Multilingual Error Annotation}\label{implementation}

Our implementation demonstrates how grammatical error annotation can be consistently extended across typologically diverse languages. We begin by reimplementing \texttt{errant} for English using \texttt{stanza} and validating its performance. We then apply the same system to other European languages without language-specific modules. For Korean, we introduce targeted refinements using language-specific templates. Finally, for Chinese, we show how deeper customization can be incorporated by modifying segmentation and retraining the \texttt{stanza} pipeline.

\subsection{Reimplementing \texttt{errant} for English}

We reimplemented \texttt{errant} using \texttt{stanza} for POS tagging and dependency parsing, as described in $\S$\ref{background:stanza}. This enables our annotation system to be more consistent across languages while preserving the linguistic precision required for English-specific grammatical labels.

We integrated the English-specific classification module from the original \texttt{errant}, which identifies detailed grammatical error types, such as \texttt{NOUN:POSS} for possessive noun suffix errors. This module relies on universal POS tags \citep{petrov-das-mcdonald:2012:LREC} and dependency relation tags to categorize errors. For instance, if the first token in an edit is tagged as \texttt{PART} and its dependency relation is \texttt{case:poss}, the classifier assigns the \texttt{NOUN:POSS} label accordingly.

A key distinction between the original \texttt{errant} and our implementation lies in error categorization. As illustrated in Figure~\ref{errant-diff}, \texttt{errant} originally annotates \textit{that is} as a missing \texttt{OTHER} error, whereas our implementation classifies it more precisely as a missing \texttt{PRON} error. Additionally, we refine verb annotation by distinguishing auxiliary verbs in passive constructions, categorizing \textit{is played} as a \texttt{R}eplacement error from \texttt{VERB} to \texttt{AUX VERB}. These refinements enhance interpretability by providing more specific and linguistically meaningful labels for complex constructions.

\begin{figure}[!ht]
\centering
\resizebox{\columnwidth}{!}{
\footnotesize{
\begin{tabular}{l}
Original \texttt{errant}: \\
\texttt{S Volleyball is a sport play every place ...}\\
\texttt{A 4 4|||M:OTHER|||that is|||REQUIRED|||-NONE-|||0} \\
\texttt{A 4 5|||R:VERB:FORM|||played|||REQUIRED|||-NONE-|||0}\\
~\\
Our implementation: \\
\texttt{S Volleyball is a sport play every place ...}\\
\texttt{A 4 4|||M:PRON|||that|||REQUIRED|||-NONE-|||0}\\
\texttt{A 4 5|||R:VERB $\rightarrow$ AUX VERB|||is played|||REQUIRED|||-NONE-|||0}\\
\end{tabular}
}
}
\caption{Differences between \texttt{errant} and our implementation}
\label{errant-diff}
\end{figure}

To evaluate the overall alignment, we compared both implementations using outputs from the state-of-the-art GEC system, T5 \citep{rothe-etal-2021-simple}. The results in Table~\ref{english-experiment-results} support that our implementation reproduces \texttt{errant}'s scores, with only minor variations.

\begin{table}[!ht]
\centering
\resizebox{\columnwidth}{!}{
\footnotesize{
\begin{tabular}{ r | cccccc} \hline 
 & TP & FP & FN & Prec & Rec & F$_{0.5}$\\ \hline  
\texttt{errant} & 2589&1639&4030&0.6123&0.3911&0.5501\\ \hdashline
{Ours} & 2565&1613&4028&0.6139&0.3890&0.5503 \\ \hline 
\end{tabular}
}
}
\caption{GEC results for English using \textsc{t5}} \label{english-experiment-results}
\end{table}

\subsection{Applying universal annotation to European languages}

\begin{figure*}
    \centering
\resizebox{\textwidth}{!}{
\footnotesize{
\begin{tabular}{r|l } \hline 
Czech     &   \texttt{S Mám velkou rodinu , tak nemohla jsem mít naději , že něco dostanu .}  \\ \hdashline
\citet{naplava-etal-2022-czech} & \texttt{A 5 7|||R:WO|||jsem nemohla|||REQUIRED|||-NONE-|||0} \\ \hdashline 
Ours & \texttt{A 5 7|||R:VERB AUX -> AUX VERB|||jsem nemohla|||REQUIRED|||-NONE-|||0}     \\
& (`I have a big family, so I couldn’t hope to get anything.')\\ \hline 
German & 
\texttt{S Dagegen wieder , bekommen BA Studenten die ein extra Jahr oder mehr studiert haben , leichter Jobs .}\\ \hdashline
\citet{boyd-2018-using} & \texttt{A 0 3|||R:OTHER|||Dahingegen|||REQUIRED|||-NONE-|||0}\\
& \texttt{A 4 5|||U:PNOUN||||||REQUIRED|||-NONE-|||0}\\
& \texttt{A 5 6|||R:NOUN|||BA-Studenten|||REQUIRED|||-NONE-|||0}\\ \hdashline
Ours 
& \texttt{A 0 2|||R:ADV ADV -> ADV|||Dahingegen|||REQUIRED|||-NONE-|||0}\\
& \texttt{A 2 3|||U:PUNCT||||||REQUIRED|||-NONE-|||0}\\
& \texttt{A 5 5|||M:PUNCT|||-|||REQUIRED|||-NONE-|||0}\\ 
&(`On the other hand, BA students who have studied an extra year or more find jobs more easily again.') \\
\hline 

\end{tabular}}}

\caption{Examples of grammatical error annotation for Czech and German} \label{european-m2-files}
\end{figure*}

Without language-specific classification modules, our grammatical error annotation system remains capable of generating generic error annotations using the core \texttt{MRU} framework combined with POS labels. We applied this approach to German \citep{boyd-2018-using} and Czech \citep{naplava-etal-2022-czech} to assess whether structured, interpretable annotations could still be produced in the absence of custom heuristics. 

As shown in Figure~\ref{european-m2-files}, our system improves clarity by attaching POS information to word order and punctuation errors, allowing more consistent cross-lingual comparisons. For Czech, the example highlights differences in word order (\texttt{WO}) annotation: our method distinguishes between auxiliary and main verbs by incorporating POS information, whereas prior work generally treated such cases as generic \texttt{WO} errors. By capturing the syntactic function of the words involved, our method enables more precise and interpretable annotation. A similar improvement is seen for German, where our universal framework avoids language-specific categories while maintaining clear and consistent labeling.


Compared to previous implementations, our annotation outputs remain broadly consistent in terms of overall operation counts, with only minor variations, as shown in Table~\ref{tab:operation_counts}. This suggests that our universal framework based on the \texttt{MRU} scheme can replicate established annotation distributions. Table~\ref{tab:top10-annotations} lists the most frequent error annotations produced by our system alongside those from previous implementations. Our system makes the syntactic categories involved in each replacement edit explicit (\texttt{R:P\textsubscript{1}~$\rightarrow$~P\textsubscript{2}}), reflecting a different annotation choice rather than a direct re-labeling of existing tags.

Future work could explore how to map between these representations to support compatibility and facilitate comparative evaluations. Another direction is to extend this annotation scheme to additional languages: because our framework leverages universal POS tags and dependency labels from \texttt{stanza}, it can be readily applied to the ten other languages in the MultiGEC dataset \citep{masciolini-etal-2025-multigec} without additional customization.

\begin{table}[!ht]
\centering
\resizebox{\columnwidth}{!}{
\footnotesize{
\begin{tabular}{ rcccc} \hline 
 & \texttt{M}issing & \texttt{R}eplacement & \texttt{U}nnecessary & Total \\ \hline  
 \textit{Czech} \\
\citet{naplava-etal-2022-czech} & 693&3707&515&4915\\ 
{Ours} & 695&3672&530&4897 \\ \hdashline
\textit{German} \\
\citet{boyd-2018-using} & 1341&4406&638&6385\\ 
{Ours} & 1310&4348&612&6270 \\ \hline
\end{tabular}
}
}
\caption{Comparison of operation counts (\texttt{M}issing, \texttt{R}eplacement, \texttt{U}nnecessary) on the development sets of Czech \citep[first 1000 sentences;][]{naplava-etal-2022-czech} and German \citep{boyd-2018-using}.} \label{tab:operation_counts}
\end{table}

\begin{table*}[!ht]
\centering
\resizebox{0.75\textwidth}{!}{
\footnotesize{
\begin{tabular}{cc|cc|cc|cc} \hline
\multicolumn{4}{c|}{\textit{Czech}} & \multicolumn{4}{c}{\textit{German}} \\
\multicolumn{2}{c}{\citet{naplava-etal-2022-czech}} & \multicolumn{2}{c|}{Ours} & \multicolumn{2}{c}{\citet{boyd-2018-using}} & \multicolumn{2}{c}{Ours} \\
Annotation & Count & Annotation & Count & Annotation & Count & Annotation & Count \\ \hline
DIACR & 989 & NOUN $\rightarrow$ NOUN & 760  & PUNCT & 942 & DET $\rightarrow$ DET & 832 \\
OTHER & 834 & VERB $\rightarrow$ VERB & 465 & SPELL & 816 & NOUN $\rightarrow$ NOUN & 814 \\
PUNCT & 487 & PUNCT & 396 & DET:FORM & 693 & PUNCT & 800 \\
SPELL & 457 & ADJ $\rightarrow$ ADJ & 299 & OTHER & 670 & ADJ $\rightarrow$ ADJ & 466 \\
VERB & 271 & PRON & 161 & ORTH & 529 & VERB $\rightarrow$ VERB & 305 \\
WO & 227 & DET $\rightarrow$ DET & 101 & ADP & 348 & DET & 241 \\
NOUN:INFL & 209 & ADV $\rightarrow$ ADV & 100 & ADJ:FORM & 277 & ADP $\rightarrow$ ADP & 170 \\
PRON & 187 & NOUN $\rightarrow$ ADJ & 98 & PRON & 273 & PRON & 170 \\
MORPH & 177 & PUNCT $\rightarrow$ PUNCT & 95 & NOUN:FORM & 260 & PRON $\rightarrow$ PRON & 144 \\
ORTH:CASING & 124 & ADP $\rightarrow$ ADP & 94 & DET & 242 & AUX $\rightarrow$ AUX & 143 \\ \hline
\end{tabular}
}
}
\caption{Comparison of the top 10 most frequent error annotations on the development sets of Czech \citep[first 1000 sentences;][]{naplava-etal-2022-czech} and German \citep{boyd-2018-using}.}
\label{tab:top10-annotations}
\end{table*}

\subsection{Refining language-specific annotations for Korean}

Previous research on Korean grammatical error annotation has relied on extensive linguistic resources \citep{yoon-etal-2023-towards}. However, grammatical errors in Korean often manifest at the morpheme level, as observed in L2 writing from the National Institute of Korean Language (NIKL) corpus. In contrast, prior error annotation approaches primarily operate at the word level, which aligns with our methodology. To ensure consistency in annotation, previous work established two priority rules for assigning a single error type to each word because of the potential ambiguity in error classification, particularly when multiple error types could apply to the same token: (i) \textsc{insertion} $>$ \textsc{deletion} $>$ others, and (ii) \textsc{ws} (word segmentation = \textsc{wb}) $>$ \textsc{wo} $>$ \textsc{spell} $>$ \textsc{shorten} (incorrect contraction of a word) $>$ \textsc{punctuation} $>$ \textsc{others}.

Building on these foundations, we implement language-specific error types based on Algorithm~\ref{error-classification-algorithm} and refine the \textsc{wb} (word boundary) category by introducing two subtypes: \textsc{wb:m} for missing spaces and \textsc{wb:u} for extraneous spaces. The former occurs when spaces are absent between words, causing multiple words to merge into a single unit, which can obscure meaning and hinder readability. The latter arises when superfluous spaces are inserted between or within words, disrupting the natural flow of the text.

Additionally, we extend grammatical error annotation to functional morphemes, categorizing errors into (i) postposition errors (\texttt{ADP}), (ii) verbal ending errors (\texttt{PART}), and (iii) honorific suffix errors (\texttt{HON}).
These errors are further classified into missing (\textsc{m}), unnecessary (\textsc{u}), and incorrect usage (\textsc{r}). 
Figure~\ref{korean-m2} illustrates corrections from two annotators: the noun phrase 음식이 \textit{eumsig-i} (`food.\textsc{nom}’) is replaced with 음식을 \textit{eumsig-eul} (`food.\textsc{acc}’), annotated as \texttt{R:NOUN -> NOUN:ADP}, reflecting a case marker correction. Similarly, 막였습니다 \textit{magyeossseubnida} is replaced with 먹었습니다 \textit{meogeossseubnida} (`ate'), which constitutes a spelling error due to phonetic and orthographic similarity.\footnote{Annotator 1 annotates a replacement with 맞았습니다 \textit{maj-assseubnida} (`agree'), altering the meaning of the sentence. This highlights a potential challenge in grammatical error annotation—distinguishing between true errors and alternative valid expressions that change sentence semantics.}

\begin{figure}[!ht]
\centering
\resizebox{.48\textwidth}{!}{
\footnotesize{
\begin{tabular}{l}
\texttt{S 비행기 $_{1}$음식이 안 $_{3}$막였습니다 .}\\
\texttt{A 1 2|||R:NOUN -> NOUN:ADP|||음식을|||REQUIRED|||-NONE-|||0}\\
\texttt{A 3 4|||R:Orthographic|||먹었습니다|||REQUIRED|||-NONE-|||0}\\
\texttt{A 3 4|||R:VERB -> VERB|||맞았습니다|||REQUIRED|||-NONE-|||1}\\
\end{tabular}
}
}
\caption{Examples from the Korean \texttt{M2} file: \textit{I didn't eat the airplane food} (Annotator 0), and \textit{The airplane food didn't agree with me} (Annotator 1)}\label{korean-m2}
\end{figure}

\subsection{Integrating deeper customization for Chinese}

\begin{figure*}[!ht]
\centering
\resizebox{0.75\textwidth}{!}{
\footnotesize{
\begin{tabular}{l}
Chinese GSD-based WB \citep{gu-etal-2025-improving}: \\
\texttt{\zh{S ... 解释 为 $_{10}$什幺 这样 的 情况 ...}}\\
\texttt{A ...}\\
\texttt{A 10 11|||R:PROPN -> PRON VERB AUX|||\zh{什么 出现 了}|||REQUIRED|||-NONE-|||0}\\
Correction: \zh{... 解释 为 什么 出现 了 这样 的 情况 ...}\\
~\\
Our LTP-based WB: \\
\texttt{\zh{S ... 解释 $_{7}$为 $_{8}$什幺 这样 的 情况 ...}}\\
\texttt{A ...}\\
\texttt{A 7 8|||R:ADP -> ADV VERB|||\zh{为什么 出现}|||REQUIRED|||-NONE-|||0}\\
\texttt{A 8 9|||R:PROPN -> AUX|||\zh{了}|||REQUIRED|||-NONE-|||0}\\
Correction: \zh{... 解释 为什么 出现 了 这样 的 情况 ...}\\
{\color{white}Correction:} \textit{... jiěshì wèishéme chūxiàn le zhèyàng de qíngkuàng ...}\\
{\color{white}Correction:} (`... explain why this kind of situation has occurred ...')
\end{tabular}
}
}
\caption{Fragments of grammatical error annotation examples in Chinese with different word boundaries. Incorrect GSD-based segmentation of \zh{为什么} \textit{wèishéme} (`why') leads to misleading annotation \zh{什么 出现 了} \textit{shénme chūxiàn le} (`what has occurred’), while LTP-based segmentation \zh{为什么 出现} \textit{wèishéme chūxiàn} (`why occurred’) provides an accurate representation.} \label{chinese-m2-files}
\end{figure*}

Chinese grammatical error annotation presents unique challenges due to the lack of explicit word boundaries \citep{qiu-etal-2025-chinese}. Previous systems \citep{zhang-etal-2022-mucgec,gu-etal-2025-improving} adopt segmentation schemes based on different linguistic assumptions: for instance, LTP\footnote{\url{https://github.com/HIT-SCIR/ltp}} emphasizes compound words as cohesive lexical units, whereas \texttt{stanza}, trained on the Chinese GSD treebank\footnote{\url{https://github.com/UniversalDependencies/UD_Chinese-GSD}}, adopts a finer-grained, morpheme-level segmentation strategy that tends to split compound expressions into smaller units.

These design choices reflect distinct philosophies rather than flaws. However, segmentation differences can affect downstream grammatical error annotation, including both the token spans and the syntactic interpretation of the correction. For example, whether a multi-character expression like \zh{为什么} \textit{wèishéme} (`why') is treated as one token or multiple (\zh{为 什么}) influences how missing or replacement errors are classified.

To illustrate the flexibility of our framework, we adopt an LTP-style segmentation approach, which aligns more closely with native speaker intuitions about lexical units in Chinese. While the default \texttt{stanza} pipeline uses GSD-style morpheme-level segmentation, our framework allows researchers to substitute this with alternative schemes, such as LTP's compound-word-based segmentation. This optional customization demonstrates that language-specific preprocessing decisions, such as tokenization granularity, can be adapted within our framework to better support accurate and interpretable error annotation.

We achieve this integration by re-annotating the Chinese GSD treebank with LTP-informed word boundaries and retraining \texttt{stanza} on this revised corpus. This ensures compatibility with our preferred segmentation standard while preserving the benefits of \texttt{stanza}'s POS tagging and parsing pipeline. As shown in Figure~\ref{chinese-m2-files}, the resulting annotations show more consistent edit spans and error categories, especially in contexts where compound expressions are frequent.

Ultimately, this customization demonstrates the modularity of our framework: rather than enforcing a one-size-fits-all solution, we allow researchers to tailor tokenization to fit linguistic expectations, making the system more robust and adaptable across languages and segmentation conventions.

\section{Conclusion}\label{conclusion}

This work advances grammatical error annotation and evaluation by introducing a standardized, modular framework for multilingual grammatical error typology. Building upon the foundations of \texttt{errant}, we designed a two-tiered system that separates language-agnostic annotation from structured language-specific extensions. This approach supports consistency across typologically diverse languages while allowing targeted customizations when needed.

We reimplemented \texttt{errant} using \texttt{stanza} to provide broader multilingual support, and demonstrated that our system produces accurate and interpretable annotations in English. We then demonstrated how our framework can be applied to other languages with varying levels of customization. For European languages, we showed that our POS- and dependency-based system can generate reliable annotations without requiring language-specific classification modules. For Korean, we applied minor refinements to capture morphologically salient features such as postpositions and spacing errors. Finally, for Chinese, we demonstrated how deeper customization--through the integration of language-specific tokenization and retraining of NLP components--can be incorporated into our framework to support fine-grained, linguistically coherent error annotation.

By balancing consistency and flexibility, our framework enables scalable, interpretable, and reusable grammatical error annotation across languages. This supports more consistent evaluation and clearer cross-linguistic comparison in multilingual GEC research.

\section*{Limitations}

While our framework presents a unified and extensible approach to multilingual grammatical error annotation, the implementations described in this paper are primarily intended to demonstrate its adaptability across different languages and levels of customization. A detailed analysis of annotation improvements, including task-specific gains and downstream evaluation effects, is left to future work.

Although we rely on existing NLP tools such as \texttt{stanza} for tokenization and parsing, which offer broad multilingual coverage and consistent annotation schemes, these tools are not explicitly optimized for processing noisy or learner-generated text. This may introduce variability in some edge cases, particularly in languages with complex morphosyntax or ambiguous word segmentation.


\appendix

\end{document}